\documentclass[runningheads]{llncs}
\usepackage{graphicx}

\usepackage[utf8]{inputenc}
\usepackage[T2A]{fontenc}
\usepackage[russian,main=english]{babel}

\begin{document}
\title{A Language Model for Grammatical Error Correction in L2 Russian}
%
%
\author{Nikita Remnev\orcidID{0000-0002-1816-3823} \and
Sergei Obiedkov\orcidID{0000-0003-1497-4001}\and
Ekaterina Rakhilina\orcidID{0000-0002-7126-0905} \and
Ivan Smirnov\orcidID{0000-0001-8361-0282} \and 
Anastasia Vyrenkova\orcidID{0000-0003-1707-7525} }

\authorrunning{N. Remnev et al.}
%
\institute{HSE University, 11 Pokrovsky Bulvar, Moscow, Russia\\\email{remnev.nikita@gmail.com, sergei.obj@gmail.com, erakhilina@hse.ru, smirnof.van@gmail.com, avyrenkova@hse.ru} }
%
\maketitle              
\begin{abstract}
Grammatical error correction is one of the fundamental tasks in Natural Language Processing. For the Russian language, most of the spellcheckers available correct typos and other simple errors with high accuracy, but often fail when faced with non-native (L2) writing, since the latter contains errors that are not typical for native speakers. In this paper, we propose a pipeline involving a language model intended for correcting errors in L2 Russian writing. The language model proposed is trained on untagged texts of the Newspaper subcorpus of the Russian National Corpus, and the quality of the model is validated against the RULEC-GEC corpus.
\end{abstract}
\section{Introduction}\label{sec:intro}

Grammatical error correction (GEC) is one of the fundamental tasks in Natural Language Processing (NLP). Errors of different types challenge GEC systems at different levels.  Most current systems are sufficiently good in dealing with typos, which can usually be corrected by considering the words that are closest to the erroneous word in terms of edit distance~\cite{Levenshtein:66,Damerau:64}. A language model can help choose the right word among several options at the same distance. More complex cases include words with several typos, agreement and lexical choice errors, and other types of errors. To address such cases, it is possible to apply manually designed rules~\cite{Sidorov:13} or use machine learning methods, in particular, approaches based on classification~\cite{Rozovskaya:19} or machine translation~\cite{Naplava:19}, which require a large amount of data (most often, tagged) for training.

One of the best spellcheckers available for the Russian language is Yandex.Speller.\footnote{\url{https://yandex.ru/dev/speller/}} This tool is capable of detecting and correcting errors in Russian, Ukrainian, and English texts by using the CatBoost machine-learning library.\footnote{\url{https://catboost.ai}} One of the advantages of Yandex.Speller is that it can handle words that are still missing from dictionaries. To correct punctuation errors specific models focused on punctuation marks are trained. Incorrect word choices and errors involving multiple words are the most difficult error types. Such errors are not only hard to correct, but they are also hard to detect. This happens because the individual items making up such expressions are written correctly; so checking for their existence in the dictionary does not help. Resorting to higher-order $N$-grams could help, but would require a critically large amount of training data.

In this paper, we focus on texts by two types of non-native (L2) speakers of Russian. The first type is people who study Russian as a foreign language. Within this group, certain words and rules used in the native language are frequently transferred into the target language. The second type is constituted by so-called heritage speakers, i.e., people who inherited Russian from their parents, but the dominant language in their environment is not Russian. For heritage speakers, typical are cases of unusual word creation and combination of two languages. Generally, L2 texts contain significantly more errors than texts by native speakers. Often, several words in a row can be misspelled, which makes it difficult to use context for error correction. Fully distorted words become even harder to recognize, since they come together with word formation patterns transferred from another language; lexical choice errors are also much more common than in native writing.


This paper is structured as follows. In Section \ref{sec:related}, we overview existing approaches to grammatical error correction. Section \ref{sec:data} describes the data we use for constructing our spellchecker and for testing it. In Section \ref{sec:model}, we describe the various steps of our approach and evaluate them experimentally. We conclude in Section \ref{sec:conclusion}. 

\section{Related Work}\label{sec:related}

Several approaches to GEC are commonly used. We will now describe these approaches in some detail.

\subsection{Rule-based approach}

The classic approach to GEC consists in designing rules for specific error types \cite{Heidorn:82,Bustamante:96}. The first rule systems were based on pattern matching substitution. Eventually, they included part-of-speech markup and information obtained from syntactic trees~\cite{Sidorov:13}. The main advantage of a rule-based system is that it can be implemented without requiring a large amount of data and complex algorithms. The drawback is that it is not possible to systematize rules that would cover all types of errors, especially for languages with rich morphology systems such as Russian. Therefore, the use of the rule-based approach has significant limitations in practice, but it can successfully complement more complex models.

\subsection{Classifier-based approach}

As the amount of annotated data has recently grown, there are now several approaches that use machine learning to train classifiers for correcting particular types of mistakes~\cite{Han:04,Rozovskaya:11}. For each type of error, a list of corrections is made, which are considered as a set of class labels. Linguistic features for the classifier may include part-of-speech tags, syntactic relationships, etc. Thus, the GEC task is solved as a multi-class classification problem in which the model selects the most appropriate candidate from the list. The classifier corrects one word for a specific category of errors, which ignores the dependencies between words in a sentence. In addition, the classifier assumes that adjacent words in a context contain no grammatical errors. However, this is not the case in texts written by non-native authors. Thus, a system that can handle only one type of errors is restricted in use. 

Over time, several methods have been developed to correct multiple errors in one sentence. One of the most commonly used approaches consists in training and combining multiple classifiers, one for each error type. In~\cite{Rozovskaya:13}, an ensemble of rule-based models and classifiers was proposed for constructing multiple correction systems. However, this approach works only if errors are independent of one another. 

Several approaches have been proposed to solve the error interaction problem. Dahlmeier ~\cite{Dahlmeier:12b} developed a beam search decoder to correct interdependent errors. This approach outperformed the ensemble of individual rule-based classifiers and models. The classification paradigm is used for Russian in~\cite{Rozovskaya:19}, where classifiers were developed for several common grammatical errors: preposition, noun case, verb aspect, and verb agreement. The experiments were conducted with training on non-native speakers' data, as well as on native speakers' data using so-called minimal supervision. The classifiers performed better in terms of the $F$-measure than the machine translation approach to which they were compared.

\subsection{Approach based on machine translation}

The current availability of a large amount of data for some languages makes it possible to design high-quality language models. In correcting grammatical errors using language models, the main idea is that sentences assigned low probability by the model are more likely to contain grammatical errors than sequences assigned a high probability.

The first works showing a successful use of language models trained on large amounts of data are~\cite{Gamon:08,Hermet:08,Yi:08}). Most of the GEC systems presented at the CoNLL 2013~\cite{Ng:13} and CoNLL 2014 \cite{Ng:14} were either based on language models or included them as components. Although, with the development of neural machine translation, approaches based solely on language models have become less popular, they continue to be an integral part of grammatical error correction modules~\cite{Junczys:16}. Alikaniotis \cite{Alikaniotis:19} propose to replace the $N$-gram language model with several implementations of modern language models of the Transformer architecture~\cite{Vaswani:17}. They evaluated their effectiveness in GEC tasks and concluded that advanced language models produce results comparable to those produced by classifier-based approaches.

In this work, we explore the potential of a language model-based approach in grammatical error correction for Russian.

\subsection{Approach based on language modeling}

The current availability of a large amount of data for some languages makes it possible to design high-quality language models. In correcting grammatical errors using language models, the main idea is that sentences assigned low probability by the model are more likely to contain grammatical errors than sequences assigned a high probability.

The first works showing a successful use of language models trained on large amounts of data are~\cite{Gamon:08,Hermet:08,Yi:08}. Most of the GEC systems presented at the CoNLL 2013~\cite{Ng:13} and CoNLL 2014~\cite{Ng:14} were either based on language models or included them as components. Although, with the development of neural machine translation, approaches based solely on language models have become less popular, they continue to be an integral part of grammatical error correction modules~\cite{Junczys:16}.~\cite{Alikaniotis:19} propose to replace the $N$-gram language model with several implementations of modern language models of the Transformer architecture~\cite{Vaswani:17}. They evaluated their effectiveness in GEC tasks and concluded that advanced language models produce results comparable to those produced by classifier-based approaches.

In this work, we explore the potential of a language model-based approach in grammatical error correction for Russian.

\section{Data}\label{sec:data}

We use the Newspaper Corpus, which is part of the Russian National Corpus (RNC)~\cite{Apresjan:06}, to build our language model. The corpus contains articles from major Russian media including \textit{Izvestia}, \textit{Komsomolskaya Pravda}, \textit{Novy Region}, \textit{RBC}, \textit{RIA Novosti}, \textit{Sovetsky Sport}, and \textit{Trud} collected from 2000 to 2011. The corpus features a diverse vocabulary and consists of a sufficiently large number of grammatically correct texts. 


To evaluate the performance of our system, we use the test sample of the RULEC-GEC corpus, which has become a benchmark for the GEC problem for the Russian language~\cite{Rozovskaya:19}. This corpus contains annotated data from the Russian Learner Corpus of Academic Writing (RULEC)~\cite{Alsufieva:12}. RULEC consists of
essays and papers written in the United States by university students learning Russian as a foreign language and heritage speakers (those who grew up in the United States but had exposure to Russian at home).

In total, the RULEC-GEC corpus includes 12480 sentences containing about 206000 words. The data was manually tagged by two annotators, who identified 13 categories of errors. 



In total, the RULEC-GEC corpus includes 12480 sentences containing about 206000 words. The data was manually tagged by two annotators, who identified 13 categories of errors. The markup in the RULEC-GEC corpus is similar to the one used  for GEC in English~\cite{Ng:14}, which allows using the MaxMatch Scorer~\cite{Dahlmeier:12a} to evaluate the results. This tool measures the performance of a system by checking how the set $e_i$ of suggested corrections for the $i$th sentence of the test set meets the gold standard correction set $g_i$ for the same sentence. Precision $P$, recall $R$, and the $F$-measure $F_{\beta}$ are defined as usual:

\begin{equation}
P = \frac{\sum^{n}_{i=1}|e_i \bigcap g_i|}{\sum^{n}_{i=1} |e_i|}
\end{equation}
\begin{equation}
R = \frac{\sum^{n}_{i=1}|e_i \bigcap g_i|}{\sum^{n}_{i=1} |g_i|}
\end{equation}
\begin{equation}
F_{\beta} = (1+\beta^2)\frac{PR}{\beta^2P + R}
\end{equation}

The MaxMatch scorer was the main scoring system for the CoNLL 2013 and CoNLL 2014 shared tasks on grammatical error correction \cite{Ng:13,Ng:14}. The $F_1$ measure was the main performance metric at CoNLL 2013 and the $F_{0.5}$ measure was the main one at CoNLL 2014. In this work, we will also use $F_{0.5}$ measure as the main performance metric.
 
It is worth noting that the texts in RULEC-GEC were written mostly by authors with a relatively high proficiency level in Russian. This is confirmed by the number of errors found in the texts contained in the corpus: it is much lower than in English corpora (Table \ref{table:3}). Some of the techniques we propose here may not be particularly suitable for correcting errors in such texts. They are more relevant for texts produced by speakers with a lower level of Russian, examples of which can be found in the Russian Learner Corpus (RLC)~\cite{Rakhilina:16}. RLC includes the data of the RULEC-GEC corpus, as well as texts of native speakers of 27 languages with different levels of proficiency in Russian. Some of the examples considered in this paper are taken from this corpus. 
\begin{table}[ht]
\caption{\centering\label{tab:other-errors} Error distribution in several corpora \cite{Rozovskaya:19}}
\label{table:3}
\centering
\begin{tabular}{ |l|r| }
\hline
\textbf{Corpus} & \textbf{Error rate} \\
\hline
Russian (RULEC-GEC) & 6.3\% \\ 
English (FCE) & 17.7\% \\ 
English (CoNLL-test) & 10.8--13.6\% \\ 
English (CoNLL-train) & 6.6\% \\ 
English (JFLEG) & 18.5--25.5\% \\ 
\hline
\end{tabular}

\end{table}

\section{Language Model for Error Correction}\label{sec:model}

In this section, we describe our approach to error correction using a language model built from (predominantly) correct texts of the Newspaper Corpus. Our approach is iterative, and we correct each sentence independently from others. Being responsible for certain error types, each stage of the algorithm starts and terminates with a number of partially corrected versions of the same sentence.

If several corrections of an error are possible, the number of versions maintained by the algorithm can grow after one iteration. This happens, in particular, when several adjacent words are misspelled, which makes it hard to choose one among several corrections based on the context. To prevent an exponential growth of versions to be maintained, we keep only a constant number (five, in the experiments reported below) of most promising ones after each iteration. These are sentences that are most likely from the point of view of the language model. We use a three-gram language model with Kneser-Key smoothing trained on the Newspaper Corpus with KenLM \cite{Heafield:13} to estimate the  sentence probability.

We preprocess the Newspaper Corpus to build a dictionary that registers the number of occurrences of each unigram and bigram and use the Symspell algorithm\footnote{\url{https://github.com/wolfgarbe/SymSpell}} to search the dictionary for words at a certain edit distance from the word to be corrected. In the following subsections, we describe the stages of our algorithm in detail.

\subsection{Correction of orthographic errors}

The first stage of our pipeline consists in correcting words that contain spelling errors. A sentence is divided into tokens, and each token is then searched in the dictionary. If the search does not return any results, the token is regarded as erroneous. For each such token, we compose a list of possible corrections based on the Damerau--Levenshtein distance and then select the most promising ones using the language model.

In L2 texts, we often encounter sequences of adjacent erroneous words. We start correcting such sequences from the rightmost word and then proceed from right to left. Although the edit distance between the incorrect and correct spellings of a word in L2 texts can be quite large, using a very large distance when searching for candidates can be prohibitive: not only it is computationally demanding, it may hopelessly complicate the selection of the right candidate among a potentially huge number of completely irrelevant ones. We found empirically that limiting the maximum distance by two (and by one for words of length at most four) at this stage yields the best results.

\begin{otherlanguage}{russian}
This however means that, for particularly distorted words, candidate lists obtained with edit distance are often empty. To address this problem, we resort to phonetic word representations yielded by the Soundex algorithm~\cite{Raghavan:04}. We build an additional dictionary where a phonetic representation is associated with a list of dictionary words that share this representation. As an example, ``пирикрасут'' is an incorrect spelling of ``перекрасят'' (`will recolor'), the edit distance between the two variants being equal to three. The phonetic representation of ``пирикрасут'', ``1090390604'', is shared by eight dictionary words including ``перекрасят''. Again, we use the Symspell algorithm to find not only words with the same phonetic representation as the erroneous word, but also words with phonetic representations at a small edit distance, in this case, at distance at most one. Among these words, we first select those at a minimum edit distance from the erroneous word, and, from these, we then choose the variants that maximize the sentence probability.
\end{otherlanguage}

The experimental results are presented in Table \ref{tab:results}. Line 1 shows the results for Yandex.Speller open-access spell checker discussed in Section \ref{sec:intro} on the RULEC-GEC data. Line 8 reports the results of the language-model approach discussed in this section. Line 15 corresponds to the results obtained by using Yandex.Speller first and then applying our approach to its output. Lines 22--27 contain the results obtained on the same data using other methods known from the literature. The other lines correspond to various improvements to be discussed in the following sections.

On its own, our approach performs worse than Yandex.Speller, especially in terms of recall. However, when used as a postprocessing step, it increases the recall, albeit at the cost of some decrease in precision, showing a higher value of the $F_{0.5}$ measure. Next, we describe several additions to our model that make it possible to further increase both precision and recall. 

\subsection{Manually designed rules}

\begin{otherlanguage}{russian}
After correcting spelling errors, we  apply two simple rules:
\begin{enumerate}
    \item Add a comma before the conjunctions ``а'' and ``что'' (but only if the previous word is not ``потому'', ``не'', or  ``ни''), as well as before forms of 
``который'' (`which').
\item When choosing between the preposition ``о'' (`about') and its form ``об'', use ``о'' when the following word starts with a consonant or a iotified vowel (``e'', ``ё'', ``ю'', ``я'') and ``об'' when followed by any other vowel. 
\end{enumerate}
\end{otherlanguage}

Although these rules admit exceptions, their application results in improved metrics on the learner corpus, see Table~\ref{tab:results}, lines 2--3, 9--10, and 16--17. This suggests that contexts presenting counterexamples to these rules rarely occur in L2 writing.

Note that the first rule concerns punctuation. Although, in general, we do not aim at correcting punctuation errors, accurate punctuation can help correct other errors at later stages, and this is why we choose to  apply this simple rule.

\subsection{Prepositions correction with RuBERT}\label{sec:rubert}

\begin{otherlanguage}{russian}
Incorrect use of prepositions is one of the most common errors made by non-native speakers. We have already discussed the incorrect usage of the prepositions ``о'' and ``об'' in the previous section. Another problematic case for non-native speakers concerns the prepositions ``в'' and ``во'' / `in' (``*во природе'' instead of ``в природе'' / `in nature'), as well as ``с'' and ``со'' / `with' (``*с своим'' instead of ``со своим'' / `with one's own'). There are also more complex cases in which a completely incorrect preposition is used; for example, ``на'' (`on') is often confused with ``в'' (`in') and vice versa (`*в рынке'' / `in the market' instead of ``на рынке'' / `on the market').
\end{otherlanguage}

To fix such errors, we apply an approach based on the neural network model RuBERT (Bidirectional Encoder Representations from Transforms (BERT)~\cite{Devlin:19} for the Russian language. To do this, we use a pretrained RuBERT for the Masked Language Modeling problem. The principle behind this operation is as follows: one token from the sentence is replaced with a masked token $<MASK>$, then the model tries to predict which token is most likely to occur in place of the mask.

To correct mistakes with prepositions, we mask all prepositions in the sentence and take the most probable option suggested by the RuBERT model. We determine whether a word is a preposition or not by using a POS-tagging solution from DeepPavlov.\footnote{\url{https://deeppavlov.ai/}} The sentence probability is calculated before and after the replacement, and, if changing the preposition sufficiently increases the probability, we accept the replacement. Since different prepositions can fit in place of the mask, it is important to set a high threshold, so as to avoid unnecessary changes in the sentence. The results obtained after this operation are presented in Table \ref{tab:results}, lines 4, 11, and 18.

\begin{otherlanguage}{russian}
The results show that a small number of errors get corrected. The corrections applied mainly concern the errors mentioned above, i.e., those in the use of the prepositions ``в'' and ``во'', as well as ``с'' and ``со''. As a result of setting a high threshold, the total number of corrected errors at this stage is small, but a lower threshold would have led to a decrease in the $F_{0.5}$ measure due to the replacement of correct prepositions. 
\end{otherlanguage}

\subsection{Correction of agreement and control errors}

Another type of errors is related to agreement and control. These errors are typical for non-native speakers, but they also occur among native speakers, although not so often. Agreement errors include subject--verb and adjective--noun agreement in gender and number. Control errors concern noun case selection depending on the governing preposition, verb, or adverb. There are other types of agreement and control errors, but, in this paper, we only consider these types as the most common in the RULEC-GEC corpus.

The POS-tagging solution from DeepPavlov mentioned in Section \ref{sec:rubert} is also used for agreement and control errors. Specifically, this solution is used to determine not only the part of speech of a word, but also the gender and number for verbs and the case, gender, and number for nouns.

\begin{otherlanguage}{russian}
To detect such errors, we extract what can be called grammatical two- and three-grams from the Newspaper Corpus. For each sentence of the corpus, we carry out POS tagging, build a parse tree, and extract all possible chains of two and three words one of which is a noun. We keep the root word and replace the dependent words with corresponding grammatical tags. For example, the sentence 
\begin{quote}\textit{В этих местах мало перспектив.}\end{quote} yields the following chains: \textit{в местах мало}, \textit{этих местах мало}, \textit{мало перспектив}. The latter chain gets transformed into\\
~\\
\textit{мало} obl | NOUN | Animacy=Inan | Case=Gen | Gender=Fem | Number=Plur
\end{otherlanguage}

When correcting errors in a sentence, we extract such two- and three-element chains from it and match them against chains extracted from the Newspaper Corpus. If some chain is not found, this is an indication of a possible error. In this case, we generate potential corrections by varying the case, number, and gender for nouns in the chain. If a resulting chain is found among the chains extracted from the Newspaper Corpus, we substitute it for the original chain in the sentence and recalculate the sentence probability. We then choose the that maximizes the increase in probability. The error correction results for each of the agreement and control error types discussed above are presented in Table \ref{tab:results}, lines 5, 12, and 19. Here again, we observe an increase in $F_{0.5}$, as well in both precision and recall.

Examining the results demonstrated by the other approaches as shown in lines 22--27 of Table \ref{tab:results}, we see that, in terms of $F_{0.5}$, our results are second to the machine-translation model from \cite{Naplava:19} with fine-tuning, which, unlike our model, needs a large amount of labeled data for training. Our model demonstrates a significantly lower recall, but its precision is slightly better than that of a fine-tuned model and much better than that of the model without fine-tuning.

\begin{table*}[ht]
\caption{\centering\label{tab:results} Results on the RULEC-GEC test sample. The best values for the precision, recall, and $F_{0.5}$ are in \textbf{bold} and the second-best values are in \textit{italics}.}
\centering
\begin{tabular}{ |c|l|c|c|c| }
\hline
\textbf{} &\textbf{Model} & \textbf{ Precision } & \textbf{ Recall } &  $ F_{0.5} $ \\
\hline
\textbf{1} & \hspace{4pt} \textbf{Yandex.Speller} & 66.17\% & 12.66\% & 35.86\% \\ 
\textbf{2} & \hspace{8pt} + Comma rules & 66.76\% & 13.53\% & 37.37\% \\ 
\hline
\textbf{3} & \hspace{8pt} + Preposition rule & 67.00\% & 13.84\% & 37.89\% \\ 
\hline
\textbf{4} & \hspace{8pt} + Preposition RuBERT & 67.23\% & 14.22\% & 38.51\% \\ 
\hline
\textbf{5} & \hspace{8pt} + Agreement and control errors & \textit{67.43\%} & 14.66\% & 39.03\% \\ 
\hline
\textbf{8} & \hspace{4pt} \textbf{Language Model} & 65.89\% & 10.16\% & 31.43\% \\ 
\textbf{9} & \hspace{8pt} + Comma rules & 66.78\% & 11.07\% & 33.29\% \\ 
\hline
\textbf{10} & \hspace{8pt} + Preposition rule & 67.15\% & 11.38\% & 33.90\% \\ 
\hline
\textbf{11} & \hspace{8pt} + Preposition RuBERT & 67.35\% & 11.75\% & 34.61\% \\ 
\hline
\textbf{12} & \hspace{8pt} + Agreement and control errors & \textbf{67.65\%} & 12.25\% & 35.35\% \\
\hline
\textbf{15} & \hspace{4pt} \textbf{Yandex.Speller + Language Model} & 63.51\% & 15.12\% & 38.73\% \\
\textbf{16} & \hspace{8pt} + Comma rules & 64.11\% & 15.99\% & 40.03\% \\ 
\hline
\textbf{17} & \hspace{8pt} + Preposition rule & 64.40\% & 16.30\% & 40.49\% \\
\hline
\textbf{18} & \hspace{8pt} + Preposition RuBERT & 64.64\% & 16.68\% & 41.03\% \\ 
\hline
\textbf{19} & \hspace{8pt} + Agreement and control errors & 64.79\% & 17.08\% & \textit{41.41}\% \\
\hline
\textbf{22} & \hspace{4pt} A. Rozovskaya, 2019~\cite{Rozovskaya:19} & 38.00\% & 7.50\% & 21.00\% \\ 
\hline
\textbf{23} & \hspace{4pt} R. Grundkiewicz, 2019~\cite{Grundkiewicz:19} & 33.30\% & \textbf{29.40\%} & 32.47\% \\
\hline
\textbf{24} & \hspace{4pt} R. Grundkiewicz, 2019~\cite{Grundkiewicz:19} with fine-tuning & 36.30\% & \textit{28.70}\% & 34.46\% \\ 
\hline
\textbf{25} & \hspace{4pt} J. Naplava, 2019~\cite{Naplava:19} & 47.76\% & 26.08\% & 40.96\% \\ 
\hline
\textbf{26} & \hspace{4pt} J. Naplava, 2019~\cite{Naplava:19} with fine-tuning & 63.26\% & 27.50\% & \textbf{50.20\%} \\
\hline
\textbf{27} & \hspace{4pt} Y. Takahashi, 2020~\cite{Takahashi:20} & 48.60\% & 16.80\% & 35.20\% \\
\hline
\end{tabular}
\end{table*}

\section{Conclusion and Future Work}\label{sec:conclusion}

In this paper, we study the usage of a language model built from correct texts for grammatical error correction in L2 writing. We use this model in isolation and for postprocessing of the results obtained with the help of another spellchecker (Yandex.Speller, in our case).  We combine it with phonetic algorithms, manually designed rules, and dedicated  procedures for specific error types, in paricular, those for prepositions, coordination and control errors. On RULEC-GEC corpus, we obtain 64.79\% precision, 17.08\% recall, and 41.41\% $F_{0.5}$, which is better than the results reported in the literature except achieved by machine-translation models with fine-tuning \cite{Naplava:19}. 

Using our language model in a postprocessing step following the application of Yandex.Speller results in increased recall, but at the cost of decreased precision. The decrease is not big though: the value of the $F_{0.5}$ measure, which favors precision over recall, is still higher with the language model than without it. Note that we obtain this improvement using a very simple language model; replacing it by a more advanced version (with better smoothing, etc.) could help avoid decrease in precision and result in an even more significant overall improvement. Our pipeline can also be extended with additional manually designed rules and procedures for specific error types.

There is room for improvement in how candidates for correcting erroneous words are generated. Currently, candidate generation is based on edit distance and phonetic representations. In L2 writing, derivational and lexical choice errors are very common. Candidate generation for those could be handled by dedicated modules based on morphological and semantic considerations.   

As future work, we also intend to incorporate into our pipeline an auxiliary part-of-speech language model that would be used together with the main language model to estimate the probability of sentences resulting from substituting various candidates for erroneous words.

It remains to see whether our approach can be tweaked to the point that it outperforms state-of-the-art models based on machine translation. It would also be interesting to see if adding our method as a postprocessing step to these models can improve their performance, as it does for Yandex.Speller.

\end{document}